\PassOptionsToPackage{table,dvipsnames}{xcolor}
\documentclass{article}

\newcommand{\articletitle}{The Sensitivity of Word Embeddings-based Author Detection Models to Semantic-preserving Adversarial Perturbations}

\newcommand{\shortarticletitle}{Semantic-preserving Perturbations and Author Detection}

\usepackage[utf8]{inputenc}
\usepackage[LFE,LAE,T1]{fontenc}
\usepackage[main=english]{babel} 

\usepackage{ifxetex}
\usepackage{hyperref}       
\usepackage{url}            
\usepackage{amsfonts}       
\usepackage{amsmath}        
\usepackage{amssymb}        
\usepackage{nicefrac}       
\usepackage{microtype}      
\usepackage{lipsum}
\usepackage{adjustbox}
\usepackage{enumitem}       
\usepackage{xcolor}         
\usepackage{comment}

\usepackage{graphicx}       
\graphicspath{{figures/}}
\DeclareGraphicsExtensions{.pdf,.jpg,.jpeg,.png}
\usepackage{multicol}
\usepackage{multirow}
\usepackage{booktabs}       
\usepackage{threeparttable} 
\usepackage[center]{caption}
\usepackage{subcaption}
\usepackage{array}          
\newcolumntype{L}[1]{>{\raggedright\arraybackslash\hspace{0pt}}m{#1}}
\newcolumntype{C}[1]{>{\centering\arraybackslash\hspace{0pt}}m{#1}}
\newcolumntype{R}[1]{>{\raggedleft\arraybackslash\hspace{0pt}}m{#1}}

\usepackage{algorithm}      
\usepackage{algpseudocode}  

\usepackage[textwidth=0.8in]{todonotes}      
\usepackage{soul}           

\usepackage[nameinlink,noabbrev]{cleveref}  

\addto\captionsenglish{}
\crefname{algorithm}{Algorithm}{Algorithms}

\usepackage{arabtex}

\ifxetex
\else
\usepackage{CJKutf8}
\fi

\usepackage{apacite}

\title{The Sensitivity of Word Embeddings-based Author Detection Models to Semantic-preserving Adversarial Perturbations}
\author{%
Jeremiah Duncan, Fabian Fallas,Chris Gropp,Emily Herron,\\
Maria Mahbub, Paula Olaya, Eduardo Ponce, Tabitha K. Samuel,\\ 
Daniel Schultz, Sudarshan Srinivasan, Maofeng Tang, Viktor Zenkov,\\ 
Quan Zhou, Edmon Begoli\footnote{All authors contributed equally.}\\
Electrical Engineering and Computer Science\\
The University of Tennessee\\
Knoxville, TN 37996\\
\texttt{jdunca51,ffallasm,cgropp,eherron5@vols.utk.edu},\\
\texttt{mmahbub,polaya,eponcemo,tsamuel@vols.utk.edu}\\
\texttt{dschult,ssriniv3,mtang4,vzenkov9@vols.utk.edu}\\
\texttt{qzhou10,ebegoli@vols.utk.edu}\\
}
\begin{document}

\maketitle

\vocalize 
\transtrue 

\setcounter{page}{1}
\thispagestyle{plain}




\section*{Abstract}
Authorship analysis is an important subject in the field of natural language processing. It allows the detection of the most likely writer of articles, news, books, or messages. This technique has multiple uses in tasks related to authorship attribution, detection of plagiarism, style analysis, sources of misinformation, etc. The focus of this paper is to explore the limitations and sensitiveness of established approaches to adversarial manipulations of inputs. To this end, and using those established techniques, we first developed an experimental framework for author detection and input perturbations. Next, we experimentally evaluated the performance of the authorship detection model to a collection of semantic-preserving adversarial perturbations of input narratives. Finally, we compare and analyze the effects of different perturbation strategies, input and model configurations, and the effects of these on the author detection model.

\section{Introduction}
Authorship identification and detection techniques are important for various natural language processing (NLP) tasks such as text mining~\cite{weber2019paratexts}, authorship attribution~\cite{burrows2007source,iqbal2013unified}, copyright infringement~\cite{perlman2019meta}, style analysis~\cite{maharjan2019jointly}, alias resolution~\cite{savoy2020elena}, co-authorship networks~\cite{ganguly2016author2vec}, user classification in social platforms~\cite{wu2020author2vec}, bot detection~\cite{dey2020detecting}, and others~\cite{zhang2017authorship}. Recently, authorship attribution has been particularly relevant in social medial and the studies of misinformation \cite{rangel2020overview,schuster2020limitations}.
Careful considerations have to be taken during authorship verification in order to circumvent a model's sensitivity to obfuscation~\cite{potthast2016author} and adversarial stylometric~\cite{brennan2012adversarial,feng2012syntactic} techniques.
It is also important to draw a distinction between the author identification and detection problems, which is largely based on the query item and selection criteria.

\paragraph{Authorship identification}
The task with authorship identification is to, given a text and a set of candidate authors, select the author that is most probable of authoring the text.
The problem of authorship identification in literary works has been investigated via multiple classification methods ranging from regression to neural networks to deep-learning~\cite{benzebouchi2019authors,ma2020towards}. The primary idea behind this identification task is to establish text characterization of consistent literary styles. Conventional machine learning techniques are successfully applied in this area, which include feature extraction, statistical computations, and feature classification~\cite{kale2017systematic}. In the machine learning task, feature engineering uses data domain knowledge for vector representation of raw texts. Authorship identification has been widely investigated based on handcrafted features, for example, syntactic~\cite{zheng2006framework}, lexical~\cite{stamatatos2013robustness}, and content-dependent~\cite{mohsen2016author} features. Mohsen et al. used variable size characters n-grams to implement author identification based on deep learning techniques, in which a stacked denoising AutoEncoder was applied to capture textual characteristics and support-vector machine is used in the classifier. Furthermore, principal components analysis and LDA (linear discriminant analysis) have been employed by~\cite{zhang2014authorship} to classify unstructured texts.

\paragraph{Authorship detection} 
The goal of authorship detection, specifically authorship verification, is to, given an author and a collection of texts, select the set of texts corresponding to the particular author. The basic approach to the development of a detector is to analyze statistical properties of text, then formulate linguistic features for detecting the unattributed text.
However, matching texts to a particular author is a challenging task because literature writing may convey a wide range of topics, text and story structure, order of events, characters, locations, themes, vocabulary, to name a few~\cite{zhao2007searching}.
Moreover, the text structure, text content, and linguistic style of a literary work created by an author can be very similar to a work created by another author~\cite{lytvyn2018linguometric}.
While there are a number of well-known authors with consistent literary styles and themes, not all authors follow patterns easy to identify, therefore, making a generalizable approach challenging.
\newline

In this work, we present a general framework for creating embedding-based classification models to solve the authorship detection problem when applied to literary works. The sensitiveness of the models is evaluated against adversarial inputs that change different stylistic aspects of the texts but not semantic ones.

\section{Approach}

The literary writers selected in this study are the three crime and detective genre novelists (Agatha Christie, Sir Arthur Conan Doyle, and Mary Roberts Rinehart), selected for the well-known, established, and understood style and structure of their works and for the availability of their novels in \textit{Project Gutenberg}~\cite{stroube2003literary}.
The approach we took is inspired by previous research (i.e.,~\cite{benzebouchi2019authors}) that uses word embeddings and a multilayer perceptron (MLP) to implement the author classification model. We have extended this approach with what we call \textit{author2vec} (see \autoref{alg:author2vec}), a general framework for constructing embedding models that capture the linguistic style of literary authors.

Initially, an author's selection of literary texts to be evaluated are compiled into a single longer text, denoted as $T$ in \autoref{alg:author2vec}.
This concatenation step does not alter the original text format.
Optional NLP pre-processing steps can then be performed, such as sentence segmentation, removal of special characters, lemmatization, and n-grams identification.
Note that stop words removal is not considered for pre-processing because we hypothesize that stop words can influence the literary style of particular authors.
The processed text is then used to create an author's embedding model. In our case, we used word2vec to create word embeddings from the author's texts.
The author embedding, $\mathcal{A}$, is a matrix of size $\lvert V\rvert \times \mathcal{H}_n$, where $V$ represents the corpus vocabulary and $\mathcal{H}_n$ the vector dimensionality.

Given that literary works tend to be large in content (i.e., typical novels contain more than 50,000 words), we represent documents as excerpts from the author's texts.
The partitioning approach used divides the concatenated text into documents represented as non-overlapping contiguous chunks of at most $N$ words each.
Then, a document vector representation is obtained by averaging the set of vectors from $\mathcal{A}$ matching the document's words, but excluding those associated with stop words.
We denote the construction of document embeddings as \textit{doc2vec} and describe the general approach in \autoref{alg:doc2vec}, where the \textit{reduce} operation collapses a set of vectors into a single vector via a reduction operation (e.g., average or sum).
In our experiments, we consider both different word embedding and document partition sizes in order to evaluate their impact on the authorship detection task and its sensitivity to perturbed inputs.

This approach, although related to Benzebouchi's model differs in two aspects, apart from using a different data cleaning pipeline. First, we have used three different MLP classifiers instead of one and broken down the multiclass classification problem into three binary classification problems. Then, instead of choosing six random representative words, we have selected 350, 1,400, or 3,500 consecutive words per document for training. The rationale behind our choices is mainly to provide more context to the classifier while training.
The following sections describe in detail the approach used to implement, parameterize, and evaluate \textit{author2vec} and the MLP classification models.

\begin{minipage}{0.95\textwidth}
    \begin{algorithm}[H]
        \small
        \renewcommand{\algorithmicrequire}{ \textbf{Input:}}
        \renewcommand{\algorithmicensure}{ \textbf{Output:}}
        \caption{author2vec: Author and document vector representations} 
        \label{alg:author2vec} 
        \begin{algorithmic}[1] 
            \Require\\
            Author corpus, $\mathcal{C}$\Comment{Vocabulary $V$}\\
            Vector embedding hyperparameters, $\mathcal{H}$\Comment{Vector dimensionality $\mathcal{H}_n$}\\
            Stopwords, $S$\\
            Document partition size, $N$
            \Ensure\\
            Author embedding, $\mathcal{A}$\Comment{$\mathcal{A} \in \mathbb{R}^{\lvert V\rvert \times \mathcal{H}_n}$}\\
            Documents embedding, $\mathcal{D}$\Comment{$\mathcal{D} \in \mathbb{R}^{\left(\lvert V\rvert - \lvert S\rvert\right) \times \left\lfloor\frac{\lvert T'\rvert}{N}\right\rfloor}$}
            \Statex
            \State $T = concatenate(\mathcal{C})$ \label{concatenate}
            \State $T' = preprocess(T)$
            \State $\mathcal{A} = word2vec(T', \mathcal{H})$
            \State $D = partition(T', N)$
            \For {$d \in D$}
                \State $k = doc2vec(d, A, S)$\Comment{$k \in \mathbb{R}^{1 \times \mathcal{H}_n}$}
                \State Add $k$ to $\mathcal{D}$
            \EndFor
        \end{algorithmic}
    \end{algorithm}
\end{minipage}

\begin{minipage}{0.95\textwidth}
    \begin{algorithm}[H]
        \small
        \caption{doc2vec: Document vector representation} 
        \label{alg:doc2vec} 
        \begin{algorithmic}[1]
        \Procedure{doc2vec}{$d$, $A$, $S$}\Comment{Document $d$, embedding $\mathcal{A}$, stop words $S$}
            \State $K = \emptyset$
            \For {$i \gets 1, \lvert d\rvert$}
                \State $w = d_i$
                \If {$w \notin S$}
                    \State Add $\mathcal{A}_w$ to $K$
                \Else
                    \State Add $[0 ... 0]$ to $K$
                \EndIf
            \EndFor
            \State $k = reduce(K)$\Comment{$K \in \mathbb{R}^{\lvert d\rvert \times \lvert\mathcal{A}_i\rvert}$}
            \State\algorithmicreturn~$k$\Comment{$k \in \mathbb{R}^{1 \times \lvert\mathcal{A}_i\rvert}$}
        \EndProcedure
        \end{algorithmic}
    \end{algorithm}
\end{minipage}

\subsection{Data Pre-processing} \label{sec:preprocess}
\begin{figure}[!htp]
    \centering
    \includegraphics[width=0.9\textwidth]{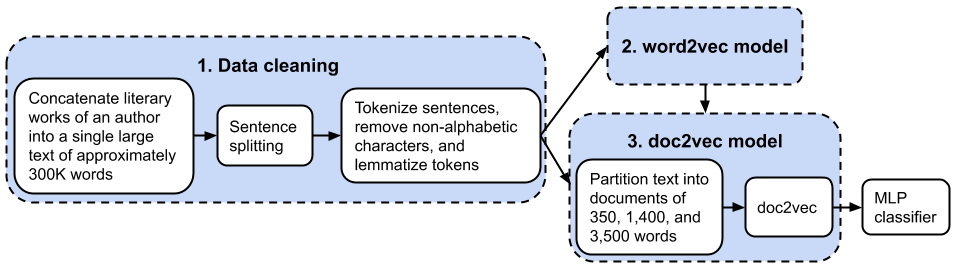}
    \caption{Pre-processing pipeline for author detection consists of data cleaning, word2vec, and doc2vec. Data cleaning pipeline includes a concatenation step, sentence segmentation, and tokenization. Document embeddings are constructed from text partitions and the word2vec model.}
    \label{fig:preprocess}
\end{figure}

Text pre-processing can play an important role in NLP tasks. It transforms text into a more precise, condensed form so that algorithms can perform better. For this work, we followed a three-step pre-processing pipeline as shown in \autoref{fig:preprocess}: data cleaning, word2vec modeling, and doc2vec modeling. The latter two steps convert discrete text data into continuous vector representations.

We performed data cleaning in three steps (\autoref{fig:preprocess}). For the sake of simplicity, we delineate the data cleaning pipeline for a single author as follows. First, we aggregate novels from the author into a single contiguous text. Then, we tokenize the larger text into sentences using the Natural Language Toolkit's (NLTK) sentence tokenizer. Each of sentence is further tokenized into words based on spaces, vertical bars, en dashes, em dashes, and periods. Finally, we remove non-alphanumeric characters using regular expressions and lemmatize verbs and nouns. The same data cleaning pipeline is applied to each author: Christie, Doyle, and Rinehart. The output of the data cleaning pipeline is used as input to the word2vec algorithm when constructing the author embedding, and fed into doc2vec when performing the classification task.

Once we have all the tokens from each author, we partition them based on data units of 350, 1,400, and 3,500 words. We have chosen these partitions to somewhat correspond to page, half-chapter and short-story structures. Each data unit partitions the author's text into multiple documents for which an embedding is formed using the doc2vec algorithm. This results in three data sets from each author, and nine data sets in total.

\subsection{Vector Embedding Modeling}

Texts are discrete data and cannot be fed into the classifier `as is'. Hence the discrete word space needs to be projected into a continuous vector space. There are numerous approaches to learning representations for words. Some approaches rely on distributional semantics and operate in a context-free manner such as word2vec~\cite{word2vec-original}, GloVe~\cite{glove-orig}, and fastText~\cite{fasttext}. These approaches are considered static embeddings. More recent approaches learn a contextual representation of words but require substantial computational power with large amounts of training data. 

Numerous static embeddings already exist and are publicly available. Many of these embeddings are trained using modern corpora that account for modern usage of words. Recent studies have shown how word-sense changes over time and across genre~\cite{hamilton-diachronic}. To account for these differences we opted to train word embeddings from scratch rather than use a word embedding from a different time period or genre. Learning word embeddings in this way allows us to capture word semantics unique to a specific author. 

\subsubsection{Author2Vec Modeling}
For this experiment, we followed the \textit{author2vec} algorithm to create word embeddings representing literary writers. We trained three independent word2vec models, one for each author considered in our study. Each model's training pipeline followed three main steps: (i)~hyperparameter optimization, (ii)~vocabulary buildup for model initialization, and (iii)~model training. After performing a systematic grid search for hyperparameter optimization, we ended up with the following hyperparameters (\autoref{table:w2v_params}) for the word2vec models:

\begin{table}[!htp]
    \setlength{\tabcolsep}{3pt}
    \centering
    \footnotesize
    \caption[word2vec parameters]{Hyperparameters of word2vec models trained on the novels by three authors.}
    \label{table:w2v_params}
    \begin{tabular}{*{9}{c}}
        \toprule
        \textbf{Author} & \textbf{algorithm} & \textbf{size} & \textbf{window} & \textbf{min\_count} & \textbf{negative} & \textbf{alpha} & \textbf{sample} & \textbf{iter}\\
        \midrule
        Christie & CBOW & 50/350 & 11 & 1 & 20  & 0.1  & 1e-3 & 7\\
        Doyle    & CBOW & 50/350 & 12 & 1 & 200 & 0.05 & 6e-5 & 5\\
        Rinehart & CBOW & 50/350 & 12 & 1 & 200 & 0.05 & 6e-5 & 5\\
        \bottomrule
    \end{tabular}
\end{table}

\begin{figure}[!htp]
     \centering
     \begin{subfigure}[t]{0.315\textwidth}
         \centering
         \fbox{\includegraphics[width=0.92\linewidth]{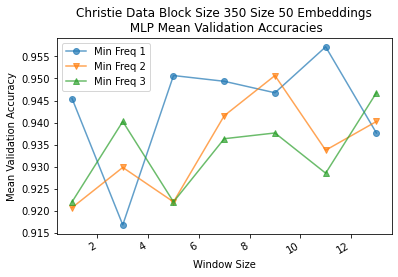}}
         \caption{50 vector dimensions, 350 data units}
         
     \end{subfigure}
     \hfill
     \begin{subfigure}[t]{0.31\textwidth}
         \centering
         \fbox{\includegraphics[width=0.92\linewidth]{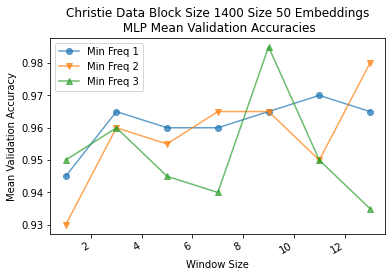}}
         \caption{50 vector dimensions, 1,400 data units}
     \end{subfigure}
     \hfill
     \begin{subfigure}[t]{0.315\textwidth}
         \centering
         \fbox{\includegraphics[width=0.92\linewidth]{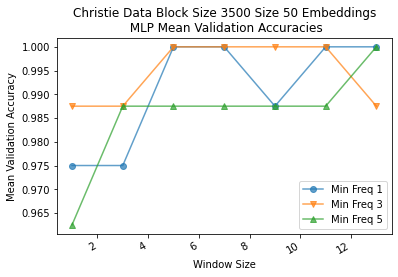}}
         \caption{50 vector dimensions, 3,500 data units}
     \end{subfigure}
     \par\bigskip
     \begin{subfigure}[t]{0.315\textwidth}
         \centering
         \fbox{\includegraphics[width=0.92\linewidth]{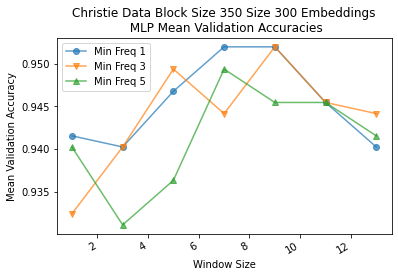}}
         \caption {300 vector dimensions, 350 data units}
     \end{subfigure}
     \hfill
     \begin{subfigure}[t]{0.315\textwidth}
         \centering
         \fbox{\includegraphics[width=0.92\linewidth]{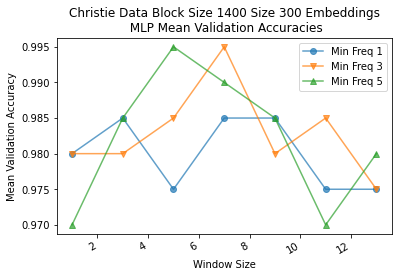}}
         \caption{300 vector dimensions, 1,400 data units}
     \end{subfigure}
     \hfill
     \begin{subfigure}[t]{0.315\textwidth}
         \centering
         \fbox{\includegraphics[width=0.92\linewidth]{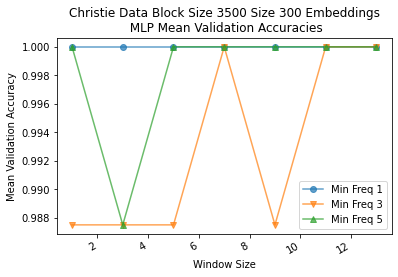}}
         \caption{300 vector dimensions, 3,500 data units}
     \end{subfigure}
     
     \caption{MLP mean validation accuracies given different minimum word frequencies and window sizes on Christie's training data.}
     \label{fig:mlp_validation_christie}
\end{figure}

\begin{figure}[!htp]
     \centering
     \begin{subfigure}[t]{0.307\textwidth}
         \centering
         \fbox{\includegraphics[width=0.94\linewidth]{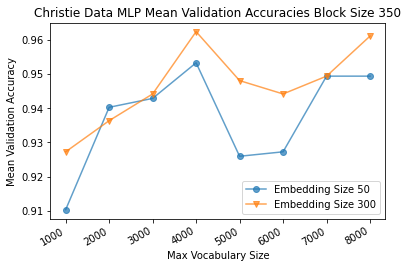}}
         \caption {350 data units}
     \end{subfigure}
     \hfill
     \begin{subfigure}[t]{0.315\textwidth}
         \centering
         \fbox{\includegraphics[width=0.94\linewidth]{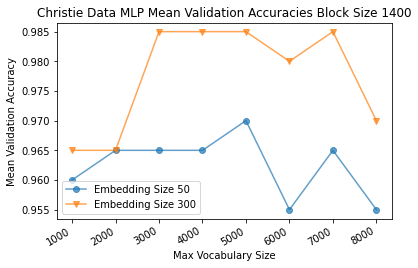}}
         \caption{1,400 data units}
     \end{subfigure}
     \hfill
     \begin{subfigure}[t]{0.315\textwidth}
         \centering
         \fbox{\includegraphics[width=0.94\linewidth]{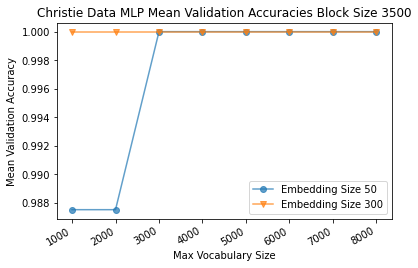}}
         \caption{3,500 data units}
     \end{subfigure}
     \caption{MLP mean validation accuracies for best fit hyperparameters given different word2vec maximum vocabulary sizes on Christie's training data.}
     \label{fig:mlp_validation1_christie}
\end{figure}

As part of the hyperparameter sweep for the word2vec models, we used validation accuracies obtained from the default MLP classifier in scikit-learn as the performance metric guide. We evaluated different minimum word frequencies and training text window sizes as well as different maximum vocabulary sizes.
We also trained the word2vec models using 50 and 300 vector dimensions, resulting in six word2vec models, two for each author.
\Cref{fig:mlp_validation_christie,fig:mlp_validation1_christie} show examples of validation accuracies resulting from the different hyperparameters experimented with, towards obtaining a best fit set of hyperparameters for the Christie data set. The vocabulary size used for all these experiments was 8,271 tokens.

The results shown in these figures suggest that the word2vec models produce better results when more information is given during training. For instance, the validation performance tended to increase with increasing maximum vocabulary sizes and number of data units. This trend was also present to a lesser extent for decreasing values of the minimum word frequency hyperparameter (which signifies the minimum number of times a word must appear in corpus to be included in the model's vocabulary). We further note that the values of the window size, minimum word frequency, and embedding size tend to become less consequential as block size increases.


\begin{figure}[!htp]
    \centering
    \begin{subfigure}[t]{0.49\textwidth}
        \centering
        \includegraphics[width=0.75\linewidth]{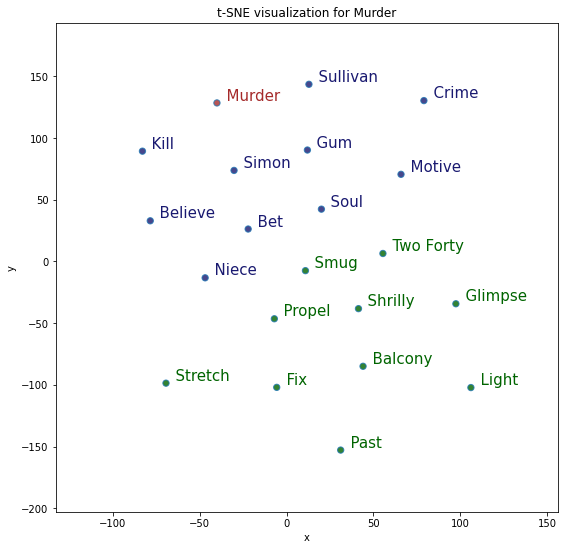}
        \caption{50 vector dimensions}
    \end{subfigure}
    \hfill
    \begin{subfigure}[t]{0.49\textwidth}
        \centering
        \includegraphics[width=0.75\linewidth]{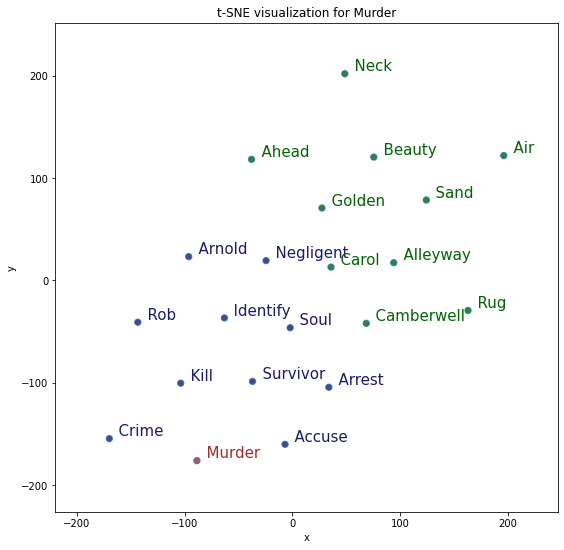}
        \caption{300 vector dimensions}
    \end{subfigure}
    
    \caption{t-SNE plot for 10 words most similar to \texttt{Murder} (in \textcolor{BlueViolet}{blue}) and 10 random words (in \textcolor{OliveGreen}{green}). Word embeddings are derived from word2vec models with different vector dimensions trained on Rinehart's novels.}
    \label{fig:model_out}
\end{figure}

\autoref{fig:model_out} presents the first two components of the t-distributed stochastic neighborhood embedding (t-SNE) for a set of words similar to \texttt{Murder} and another set of random words. These word embeddings correspond to the word2vec models trained from Rinehart's novels. Note that for both embedding sizes the surrounding words, such as \texttt{Kill}, \texttt{Motive}, \texttt{Crime}, \texttt{Survivor}, clearly have close relation to the word \texttt{Murder}. This shows the strength of the word-embedding space, that is, the author embedding.

\subsubsection{Doc2Vec Modeling}
\begin{figure}[!htp]
     \centering
     \begin{subfigure}[t]{0.315\textwidth}
         \centering
         \fbox{\includegraphics[width=0.92\linewidth]{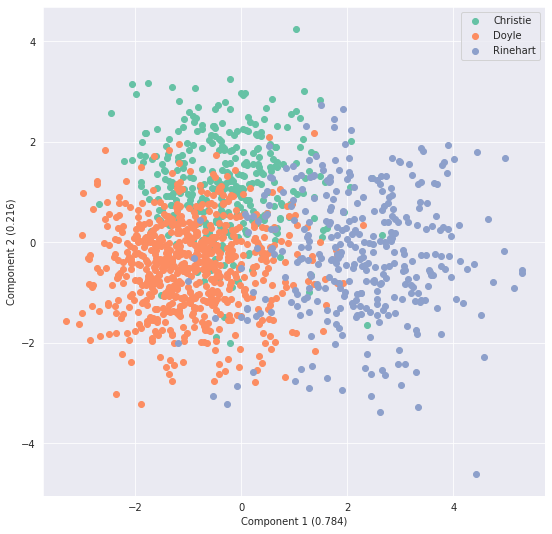}}
         \caption {50 vector dimensions, 350 data units}
     \end{subfigure}
     \hfill
     \begin{subfigure}[t]{0.315\textwidth}
         \centering
         \fbox{\includegraphics[width=0.92\linewidth]{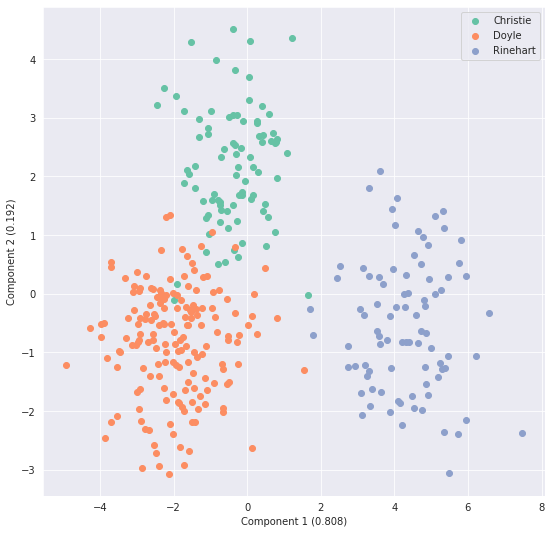}}
         \caption{50 vector dimensions, 1,400 data units}
     \end{subfigure}
     \hfill
     \begin{subfigure}[t]{0.315\textwidth}
         \centering
         \fbox{\includegraphics[width=0.92\linewidth]{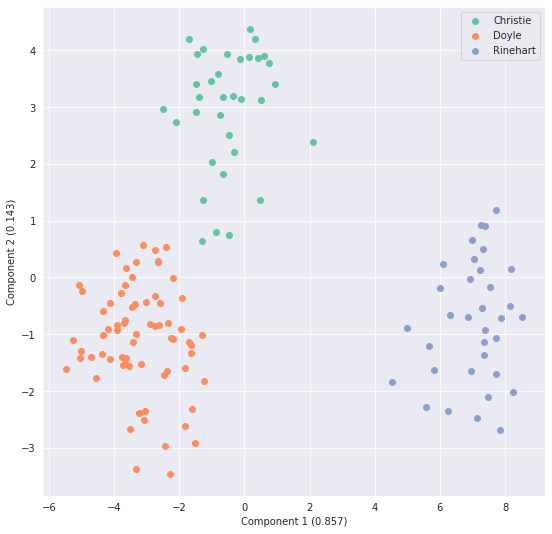}}
         \caption{50 vector dimensions, 3,500 data units}
     \end{subfigure}
     \par\bigskip
     \begin{subfigure}[t]{0.31\textwidth}
         \centering
         \fbox{\includegraphics[width=0.92\linewidth]{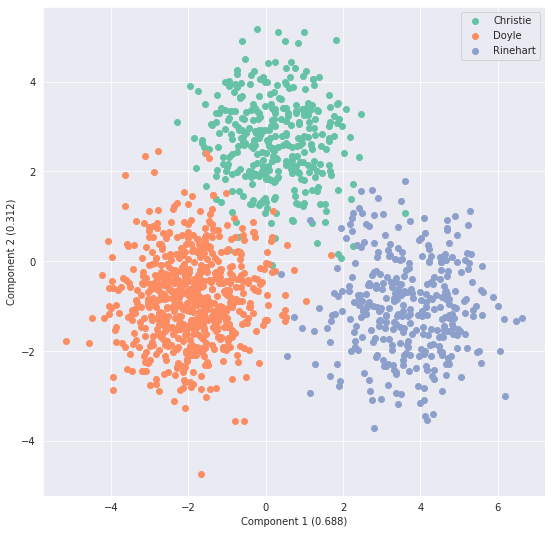}}
         \caption {300 vector dimensions, 350 data units}
     \end{subfigure}
     \hfill
     \begin{subfigure}[t]{0.315\textwidth}
         \centering
         \fbox{\includegraphics[width=0.92\linewidth]{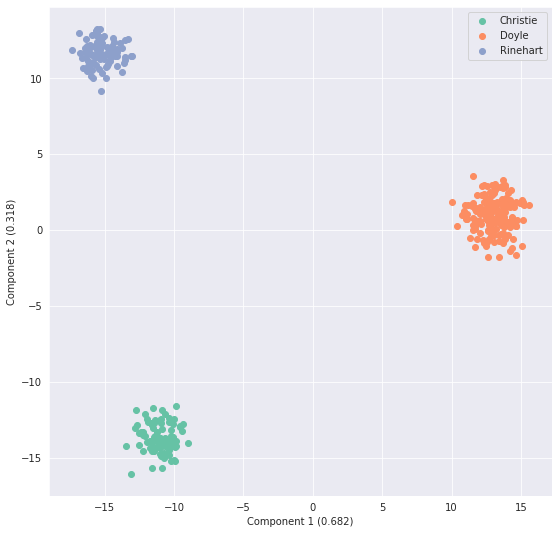}}
         \caption{300 vector dimensions, 1,400 data units}
     \end{subfigure}
     \hfill
     \begin{subfigure}[t]{0.315\textwidth}
         \centering
         \fbox{\includegraphics[width=0.92\linewidth]{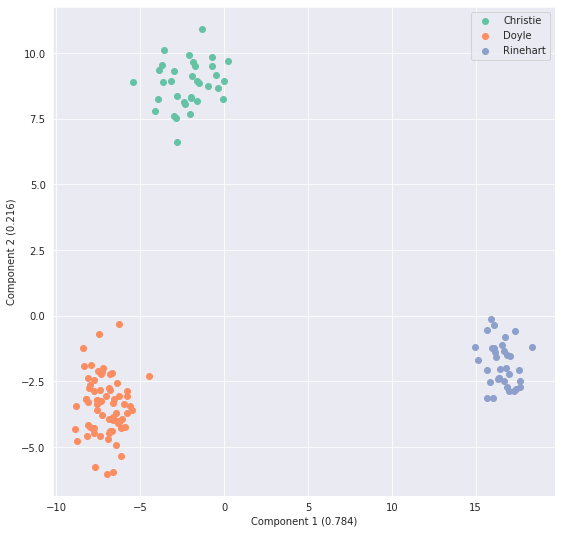}}
         \caption{300 vector dimensions, 3,500 data units}
     \end{subfigure}
     
     \caption{Linear discriminant analysis of document embeddings on authors' training data. The first two components capture over 90\% of the variance. Larger embedding size and document partition sizes produce author-specific clusters corresponding to differences in authors' literary styles.}
     \label{fig:doc2vec_doyle}
\end{figure}

The inputs to the MLP classification model are document embeddings constructed from the author embedding using documents of three different sizes. We evaluated document embeddings using LDA and projected their first two components, see \autoref{fig:doc2vec_doyle}. The visualization clearly shows that documents for a particular author cluster and, in most cases, do not overlap with clusters pertaining to another author's documents. The document embeddings are able to capture more differences between authors for embedding size 300 and document partition sizes of 1,400 and 3,500 words. The distinguishable clusters of documents is attributed to inherent differences between authors' linguistic styles and vocabulary. 

\subsection{Classification Modeling}

For the classification step, we implemented the traditional multilayer perceptron~\cite{popescu2009multilayer} model. This neural network architecture has been widely used for text classification tasks, refer to~\cite{gururangan2019variational,sajnani2011multi,SRIVASTAVA2020319}. In particular, we used the \textit{MLPClassifier} model provided by the scikit-learn library. \autoref{tab:classifierParams2} shows the author-specific set of hyperparameters selected using a grid search with cross-validation. 

\begin{table}[!htp]
    \setlength{\tabcolsep}{3pt}
    \centering
    \footnotesize
    \caption[MLP parameters]{Hyperparameters of MLP models trained on the novels by three authors.}
    \label{table:mlp_params}
    \begin{tabular}{*{6}{c}}
        \toprule
        \textbf{Author} & \textbf{hidden\_layer\_sizes} & \textbf{solver} & \textbf{alpha} & \textbf{max\_iter} & \textbf{tol}\\
        \midrule
        Christie & 25       & Adam & 1e-3 & 3000 & 1e-4\\
        Doyle    & 100      & Adam & 1e-4 & 2000 & 1e-9\\
        Rinehart & 50,50,50 & SGD  & 1e-4 & 2000 & 1e-9\\
        \bottomrule
    \end{tabular}
\label{tab:classifierParams2}
\end{table}

It is interesting to note here that, a single set of hyperparameters did not led to similar validation accuracies across authors, even though the models themselves were trained on the specific author data set. This suggests that there is close alignment between an author's writing style and the hyperparameters that work best at identifying that style.

\subsection{Perturbation of Data} 
The second focus of this paper was to create adversarial inputs based on the authors' texts. The goal here was to attempt to deceive the models trained with each author, and evaluate their performance sensitivity to perturbed variants of the original authors' texts. Perturbations were done in such a manner as to completely retain the semantics of the text, while limiting the number of words replaced to 20\% from the total word count in the corpus. For this task, we applied four different perturbation techniques to create adversarial inputs: (i)~synonym replacement was applied to all authors' testing data set, (ii)~contractions were expanded for Christie's testing data set, (iii)~translation between American and British English was applied to Rinehart's testing data set, and (iv)~numerical transformations, where literal numbers were changed to their textual form, were applied to Doyle's testing data set. 

\subsubsection{Synonym Replacement}
\begin{figure}[!htp]
     \centering
     \begin{subfigure}[t]{0.32\textwidth}
         \centering
         \fbox{\includegraphics[width=0.92\linewidth]{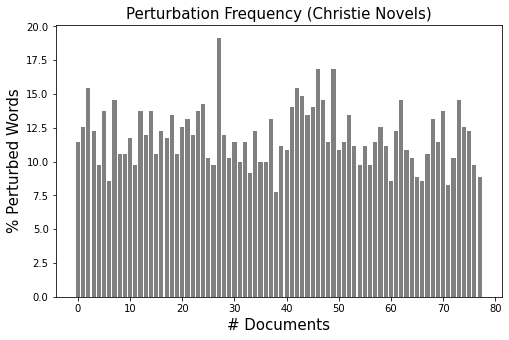}}
         \caption {350 data units}
     \end{subfigure}
     \hfill
     \begin{subfigure}[t]{0.315\textwidth}
         \centering
         \fbox{\includegraphics[width=0.92\linewidth]{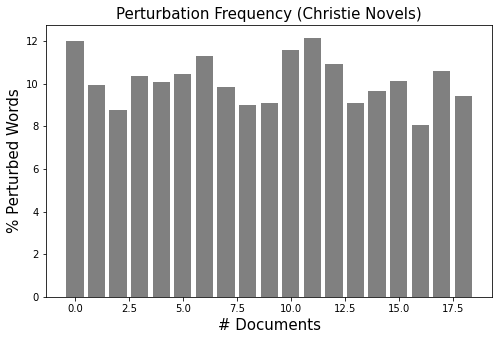}}
         \caption{1,400 data units}
     \end{subfigure}
     \hfill
     \begin{subfigure}[t]{0.315\textwidth}
         \centering
         \fbox{\includegraphics[width=0.92\linewidth]{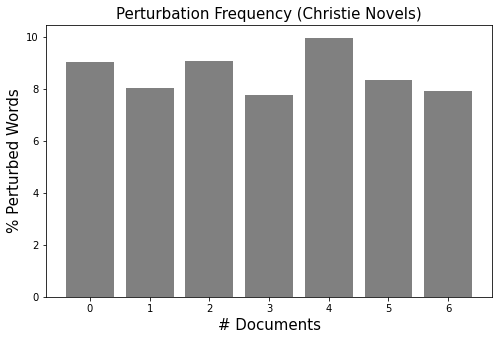}}
         \caption{3,500 data units}
     \end{subfigure}
     \caption{Percentage of synonym replacement perturbation on Christie's testing data set}
     \label{fig:syn_repl_christie}
\end{figure}

In this approach, we considered only singular nouns, singular present form of verbs, adjectives, and adverbs to make sure that the context, grammar, tense, meaning, and the overall linguistic quality of the novels were preserved. We also excluded words with less than four characters. From these words, we find synonym candidates using NLTK's \textit{Synsets}. However, these sets are quite broad, and include word usages that may be irrelevant in our context; for example, the word \texttt{sentence} can mean ``phrase'' or ``judgment'' but the latter is much more likely to preserve meaning in a crime novel. As such, we evaluated these candidates using the word embeddings trained previously, removing candidates with cosine similarities to the original word below 0.2. We also carefully discarded synonyms with similarity score 1.0, since this indicated the candidate was not different from the original token. We also removed candidates whose part-of-speech (POS) differed from the original, using NLTK's POS tagger. If no candidates remained, we left the word unperturbed. \autoref{fig:syn_repl_christie} shows the percentage of synonym replacement perturbation for each document in Christie's testing data set. The perturbation ratio is consistent across documents of the same size, and none reached the 20\% threshold. Similar results were obtained for the other authors.

The following is a sample of Christie's novel before and after applying synonym-based perturbations. 

\paragraph{Original text:} The \_Lusitania\_ had been struck by two torpedoes in \textcolor{ForestGreen}{succession} and was sinking rapidly, while the boats were being launched with all \textcolor{ForestGreen}{possible} speed. The women and children were being lined up awaiting their turn. Some still clung \textcolor{ForestGreen}{desperately} to husbands and fathers; others clutched their children \textcolor{ForestGreen}{closely} to their breasts. One girl stood alone, \textcolor{ForestGreen}{slightly} \textcolor{ForestGreen}{apart} from the rest. She was \textcolor{ForestGreen}{quite} young, not more than eighteen. She did not \textcolor{ForestGreen}{seem} afraid, and her grave, \textcolor{ForestGreen}{steadfast} eyes looked straight ahead.

\paragraph{Perturbed text:} The \_Lusitania\_ had been struck by two torpedoes in \textcolor{Bittersweet}{sequence} and was sinking rapidly, while the boats were being launched with all \textcolor{Bittersweet}{potential} speed. The women and children were being lined up awaiting their turn. Some still clung \textcolor{Bittersweet}{urgently} to husbands and fathers; others clutched their children \textcolor{Bittersweet}{intimately} to their breasts. One girl stood alone, \textcolor{Bittersweet}{somewhat} \textcolor{Bittersweet}{aside} from the rest. She was \textcolor{Bittersweet}{rather} young, not more than eighteen. She did not \textcolor{Bittersweet}{appear} afraid, and her grave, \textcolor{Bittersweet}{unwavering} eyes looked straight ahead.

\subsubsection{Contraction Expansion and Changing First-person Pronouns}
\begin{figure}[!htp]
     \centering
     \begin{subfigure}[t]{0.32\textwidth}
         \centering
         \fbox{\includegraphics[width=0.92\linewidth]{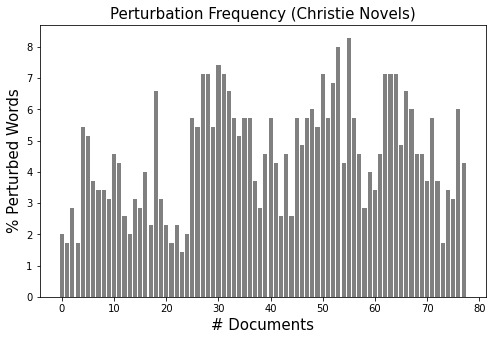}}
         \caption {350 data units}
     \end{subfigure}
     \hfill
     \begin{subfigure}[t]{0.315\textwidth}
         \centering
         \fbox{\includegraphics[width=0.92\linewidth]{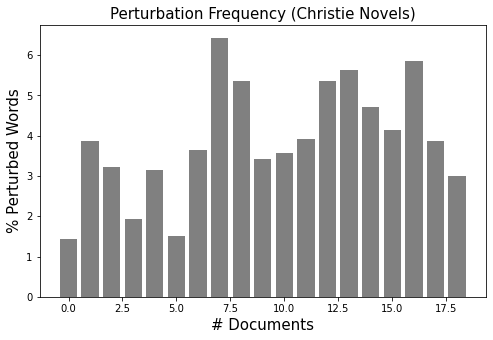}}
         \caption{1,400 data units}
     \end{subfigure}
     \hfill
     \begin{subfigure}[t]{0.315\textwidth}
         \centering
         \fbox{\includegraphics[width=0.92\linewidth]{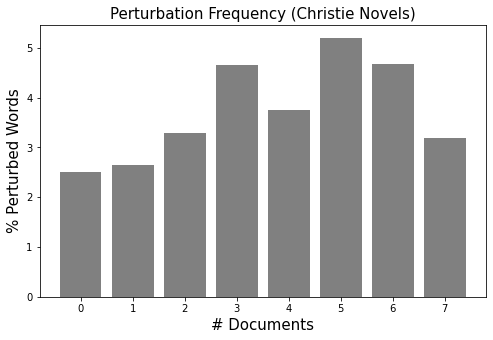}}
         \caption{3,500 data units}
     \end{subfigure}
     \caption{Percentage of contraction expansion and change of first-person pronoun perturbations on Christie's testing data set}
     \label{fig:con_pronouns_christie}
\end{figure}

Contractions may affect the performance of NLP models if these are not able to discern the equivalence between contractions and their expanded counterparts. Also, expanding contractions can increase the vocabulary size. Our pre-processing pipeline retains stop words, so contractions can be used as adversarial instruments. Moreover, pronouns in crime novels are a key characteristic because many of them have first-person narratives.
To explore the effects of contractions and first-person pronoun when detecting an author, we expanded all contractions and replaced all instances of \texttt{I} with \texttt{myself}. \autoref{fig:con_pronouns_christie} shows the percentage of these perturbations for each document in Christie's testing data set.

The following is a sample of Christie's novel before and after applying the contractions and first-person pronoun perturbations.

\paragraph{Original text:} \ldots\textcolor{ForestGreen}{you’ll} have some question to answer, and after \textcolor{ForestGreen}{you’ve} answer them we shall know what to do with you and \textcolor{ForestGreen}{I} can tell you young lady\ldots

\paragraph{Perturbed text:} \ldots\textcolor{Bittersweet}{you will} have some question to answer, and after \textcolor{Bittersweet}{you have} answer them we shall know what to do with you and \textcolor{Bittersweet}{myself} can tell you young lady\ldots

\subsubsection{American/British English Language Translation}
\begin{figure}[!htp]
     \centering
     \begin{subfigure}[t]{0.315\textwidth}
         \centering
         \fbox{\includegraphics[width=0.92\linewidth]{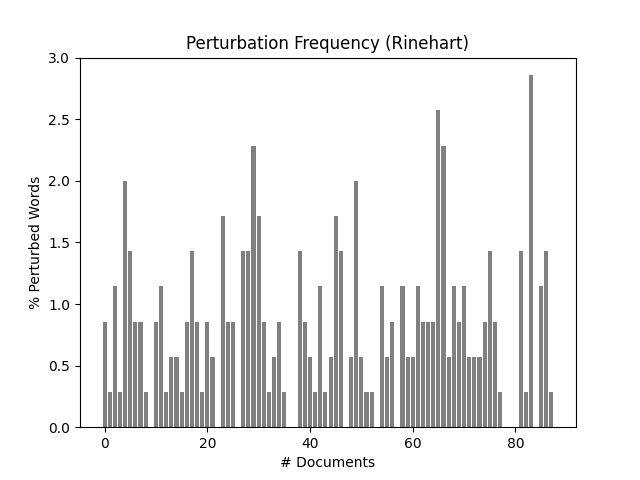}}
         \caption {350 data units}
     \end{subfigure}
     \hfill
     \begin{subfigure}[t]{0.315\textwidth}
         \centering
         \fbox{\includegraphics[width=0.92\linewidth]{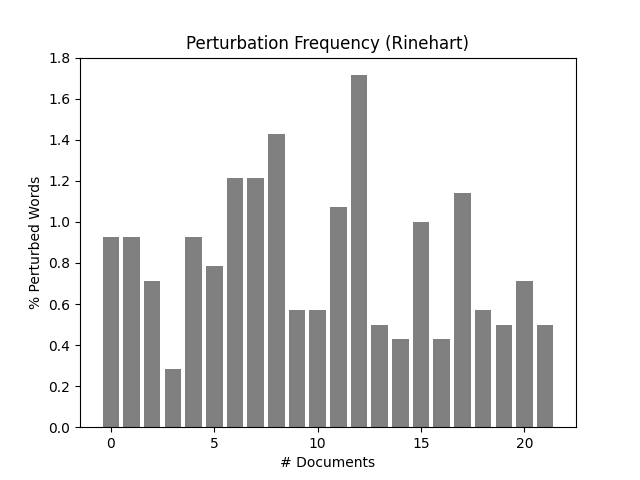}}
         \caption{1,400 data units}
     \end{subfigure}
     \hfill
     \begin{subfigure}[t]{0.315\textwidth}
         \centering
         \fbox{\includegraphics[width=0.92\linewidth]{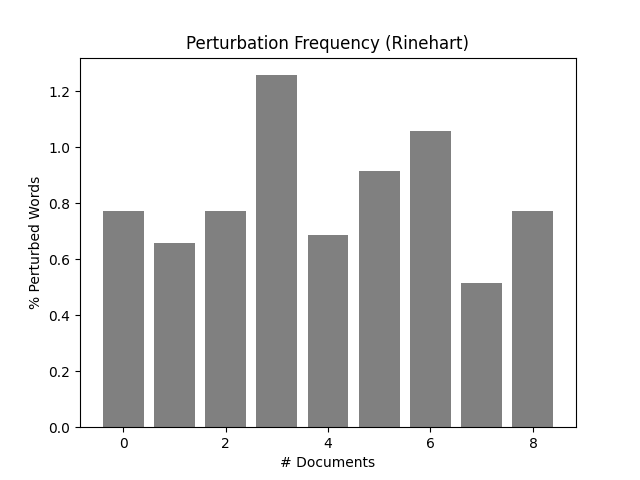}}
         \caption{3,500 data units}
     \end{subfigure}
     \caption{Percentage of American to British English translation on Rinehart's testing data set}
     \label{fig:langtrans_repl_rinehart}
\end{figure}

Given that each author uses distinct English dialect, we wanted to attack the classification models by changing the dialect each author uses. Specifically, we decided on converting American English to British English and vice versa. The direction of the language translation was selected to be opposite of the author's original language. For example, Rinehart's text was translated to British English while American English was used for the other authors. We collected a list of common words that have the same meaning but different spellings from the \textit{Comprehensive list of American and British spelling differences}\footnote{\url{http://www.tysto.com/uk-us-spelling-list.html}}. Each document was perturbed by replacing all instances of spellings from one dialect to the other. \autoref{fig:langtrans_repl_rinehart} shows the percentage of language translation applied to each document in Rinehart's testing data set.

The following is a sample of Rinehart's novel before and after applying the English dialect-based perturbation. 

\paragraph{Original text:} ``That is one, and the other is, `If you get too noisy, and the patrol \textcolor{ForestGreen}{wagon} comes, make the driver take you home.'\,'' The crowd was good-\textcolor{ForestGreen}{humored};

\paragraph{Perturbed text:} ``That is one, and the other is, `If you get too noisy, and the patrol \textcolor{Bittersweet}{waggon} comes, make the driver take you home.'\,'' The crowd was good-\textcolor{Bittersweet}{humoured};

\subsubsection{Numerical Transformations}
In numerical perturbations, text was altered such that all numbers are transformed into a word format (e.g., \texttt{3} is converted to \texttt{three}).
This was achieved using a combination of regular expressions and the Python library \textit{Inflect}\footnote{\url{https://pypi.org/project/inflect}}. After tokenization, regular expressions are used to find the numerical tokens in the text and then, these are handed off to \textit{Inflect} to apply the transformation.
As a result of the token expansion, the vocabulary size can increase.
The Doyle's testing data set was comprised of 27,929 tokens containing 13 numerical tokens. These perturbations represent an alteration of approximately 0.05\% of the text. In other words, this is a limited adversarial attack.

The following is a sample of Doyle's novel before and after applying the numerical perturbation.

\paragraph{Original text:} Over the low, heavily-lintelled door, in the centre of this old part, is chiseled the date, \textcolor{ForestGreen}{1607}, but experts are agreed that the beams and stonework are really much older than this.

\paragraph{Perturbed text:} Over the low, heavily-lintelled door, in the centre of this old part, is chiseled the date, \textcolor{Bittersweet}{one thousand, six hundred and seven}, but experts are agreed that the beams and stonework are really much older than this.

\section{Experiments}

In this section we describe the setup of the experiments, selected corpora, and the considerations for splitting into training, validation, and testing data sets in our work.
We also expose the tools used for data processing, models construction, and visualizations.

\subsection{Authors and Data}
We selected several novels (and short stories) from three well-known mystery authors -- Christie, Doyle, and Rinehart (\autoref{tab:corpus_structure}), with each data set being set to be approximately 300,000 words. Novels were chosen such that they adhere to the \textit{whodunit} paradigm~\cite{herbert2003whodunit}. The total number of words and characters is similar among authors, with less than 10\% difference, but in terms of sentence count, Doyle has notably less, indicative that his style either makes use of longer sentences or infrequently uses symbols searched by the sentence tokenizer. The number of documents produced by each partition size is consistent across authors as observed in \autoref{tab:corpus_documents}.

The data set for each author was split into 90\% for training each MLP model, and the other 10\% for testing. From the training data set 90\% was selected for training and 10\% for validation. We applied the same ratio split for hyperparameter tuning of the MLP models.
Moreover, we have two scenarios from the testing data sets: (i)~original data and (ii)~perturbed data. Perturbations were applied either commonly across authors or in an author-specific manner. The synonym replacement perturbation was evaluated for all three author data sets. The three author-specific perturbations chosen were transforming contractions and first-person pronouns for Christie's corpus, language translation (American/British English) for Rinehart's corpus, and converting literal numbers to text form for Doyle's corpus.

\begin{table}[!htp]
    \setlength{\tabcolsep}{3pt}
    \centering
    \footnotesize
    \caption[Text structure]{Structural decomposition of all author's novels/stories.}
    \label{tab:corpus_structure}
    \begin{threeparttable}
        \begin{tabular}{c L{2in} ccc}
            \toprule
            \textbf{Author} & \multicolumn{1}{c}{\textbf{Novel}} & \textbf{Sentences} & \textbf{Words} & \textbf{Characters}\\
            \midrule
            \multirow{4}{*}{Christie}
                & The Man in the Brown Suit & 7,064 & 75,924 & 421,547 \\
                & The Secret Adversary & 8,347 & 75,869 & 435,123 \\
                & The Murder on the Links & 6,294 & 64,981 & 371,313 \\
                & The Mysterious Affair at Styles & 5,770 & 57,096 & 328,929 \\
            \cmidrule{3-5}
            & & \textbf{27,475} & \textbf{273,870} & \textbf{1,556,912} \\
            \midrule
            \multirow{12}{*}{Doyle}
               & The Hound of the Baskervilles & 4,074 & 59,413 & 325,762 \\
               & The Valley of Fear & 4,465 & 57,884 & 318,639 \\
               & A Study in Scarlet & 2,725 & 43,501 & 243,059 \\
               & The Sign of the Four & 3,020 & 43,435 & 237,222 \\
               & The Adventure of the Speckled Band\tnote{1} & 645 & 9,887 & 54,145 \\
               & The Adventure of the Second Stain\tnote{1} & 806 & 9,736 & 54,316 \\
               & The Adventure of the Dancing Men\tnote{1} & 647 & 9,717 & 52,710 \\
               & The Boscombe Valley Mystery\tnote{1} & 663 & 9,685 & 52,449 \\
               & The Adventure of the Cardboard Box\tnote{1} & 526 & 8,743 & 51,760 \\
               & The Musgave Ritual\tnote{1} & 405 & 7,620 & 45,713 \\
               & The Five Orange Pips\tnote{1} & 474 & 7,367 & 40,342 \\
               & The Reigate Squires\tnote{1} & 504 & 7,227 & 44,547 \\
            \cmidrule{3-5}
            & & \textbf{18,954} & \textbf{274,215} & \textbf{1,520,664} \\
            \midrule
            \multirow{4}{*}[-4pt]{Rinehart}
                & The Circular Staircase & 5,270 & 70,631 & 391,008 \\
                & The Window at the White Cat & 5,000 & 68,898 & 378,221 \\
                & The Man in Lower Ten & 5,144 & 64,940 & 359,520 \\
                & The After House & 4,152 & 48,280 & 269,882 \\
                & The Case of Jennie Brice & 3,402 & 36,096 & 193,950 \\
            \cmidrule{3-5}
            & & \textbf{22,968} & \textbf{288,845} & \textbf{1,592,581} \\
            \bottomrule
        \end{tabular}
        \begin{tablenotes}
            \item[1] Short story
        \end{tablenotes}
    \end{threeparttable}
\end{table}

\begin{table}[!htp]
    \setlength{\tabcolsep}{3pt}
    \centering
    \footnotesize
    \caption[Document count]{Document count for each author and partition size.}
    \label{tab:corpus_documents}
    \begin{tabular}{C{0.5in} *{3}{C{0.4in}}}
        \toprule
        \multirow{2}{*}[-2.5pt]{\centering\textbf{Author}}
        & \multicolumn{3}{c}{\textbf{Document Data Units}}\\
            \cmidrule(lr){2-4}
            & \textbf{350} & \textbf{1,400} & \textbf{3,500}\\
        \midrule
        Christie & 766 & 194 & 78 \\
        Doyle    & 760 & 194 & 78 \\
        Rinehart & 803 & 204 & 82 \\
        \bottomrule
    \end{tabular}
\end{table}

\subsection{Setup and Experimental Environment}
All the experiments were performed in Google Colaboratory notebooks to ensure an organized, reproducible, and shareable project workflow. We used a variety of Python libraries: NumPy, pandas for data analysis, NLTK and spaCy for data cleaning and perturbations, regular expression module (regex), Gensim for word2vec modeling, scikit-learn for classification models, and Matplotlib and Seaborn for data visualization.

\section{Results}

We present the results for our two testing scenarios: (i)~original data, and (ii)~perturbed data. We used accuracy, the proportion of true results among the total number of cases examined, as our evaluation metric. Results corresponding to adversarial inputs are used to establish sensitivity analyses for the classification models.

\subsection{Performance on Original Data}
In \autoref{tab:classifier_accuracy_perturbed} we present the accuracy for each word embedding size and document partition size. For Christie's data we found that the highest accuracy of 0.98 is obtained with 50 as the embedding size and 1,400 as the data unit. Overall, Doyle and Rinehart models achieved the highest accuracy of 1.0 with the largest data unit (3,500) and both embedding sizes.
In general, the larger the data unit, meaning more words per document, and a small embedding size (50), capture more relevant information allowing higher author prediction accuracy by the MLP model.
We infer that the higher accuracies are indicative of differences between author styles which are recognized by the classification models.

\subsection{Performance on Adversarial Modified Inputs}
The first finding as observed in \autoref{tab:classifier_accuracy_perturbed} is that when the synonym perturbation is applied to the testing data, an accuracy of 1.0 is obtained with the largest vector embedding size (300) and the largest document partition size (3,500) for all authors.
An interesting observation here is that for both the original and synonym perturbations data set, a high accuracy (over 95\%) was achieved across all authors corpora, even though the synonym perturbation ratio was highest among perturbation types.
This leads us to the conclusion that under the right configuration of hyperparameters, a model can be chosen that can interpret the structure of the text well enough, and that changing the words, but not the meaning of the sentence does not have an impact on the performance of the algorithms.
Another interesting observation is that the smallest document size of 350 words performed the best for the contractions perturbations. Larger word embedding size of 300 performed better than the smaller embedding size of 50 across all perturbations.
The perturbation type that had the least adverse impact was the English style translation, regardless of embedding and document size.
Note that the impact of the perturbation to increase the accuracy compared to the one obtained with the original data is not covered in this paper. 

\begin{table}[!htp]
    \setlength{\tabcolsep}{3pt}
    \centering
    \footnotesize
    \caption[Classification accuracy on perturbed text]{Accuracy of classification task on author's perturbed text.}
    \label{tab:classifier_accuracy_perturbed}
    \begin{threeparttable}
    \begin{tabular}{*{4}{C{0.5in}} *{2}{C{0.52in}} *{2}{C{0.5in}}}
        \toprule
        \multirow{2}{0.5in}[-7pt]{\centering\textbf{Author}}
        & \multirow{2}{0.5in}[-7pt]{\centering\textbf{Vector Size}}
        & \multirow{2}{0.5in}[-7pt]{\centering\textbf{Document Size}}
        & \multirow{2}{0.5in}[-7pt]{\centering\textbf{Original Text}}
        & \multicolumn{4}{c}{\textbf{Perturbation Type}\tnote{1}}\\
            \cmidrule(lr){5-8}
            & & & & \textbf{Synonym Replace} & \textbf{US/UK Translate} & \textbf{Cntrac'n Pronoun}\tnote{2} & \textbf{Number to Text}\\
         \midrule
        \multirow{6}{*}[-1pt]{Christie}
        & \multirow{3}{*}{50}
          &   350  & 0.90 & 0.85 & & 0.88 & \\
          & & 1400 & 0.98 & 0.95 & & 0.79 & \\
          & & 3500 & 0.69 & 1.00 & & 0.71 & \\
          \cmidrule{2-8}
        & \multirow{3}{*}{300}
          &   350  & 0.85 & 0.94 & & 0.96 & \\
          & & 1400 & 0.96 & 1.00 & & 0.94 & \\
          & & 3500 & 0.95 & \color{red}{1.00} & & 0.95 & \\    
        \midrule
        \multirow{6}{*}[-1pt]{Doyle}
        & \multirow{3}{*}{50}
          &   350  & 0.83 & 0.82 & & & 0.89 \\
          & & 1400 & 0.95 & 0.89 & & & 0.89 \\
          & & 3500 & 1.00 & 0.95 & & & 0.98 \\
          \cmidrule{2-8}
        & \multirow{3}{*}{300}
          &   350  & 0.88 & 0.88 & & & 0.91 \\
          & & 1400 & 0.98 & 0.95 & & & 0.95 \\
          & & 3500 & 1.00 & \color{red}{0.96} & & & 1.00 \\
        \midrule
        \multirow{6}{*}[-1pt]{Rinehart}
        & \multirow{3}{*}{50}
          &   350  & 0.95 & 0.95 & 0.92 & & \\
          & & 1400 & 0.98 & 0.97 & 0.97 & & \\
          & & 3500 & 1.00 & 1.00 & 0.96 & & \\
          \cmidrule{2-8}
        & \multirow{3}{*}{300}
          &   350  & 0.93 & 0.92 & 0.91 & & \\
          & & 1400 & 0.98 & 0.98 & 0.98 & & \\
          & & 3500 & 1.00 & \color{red}{1.00} & 1.00 & & \\
       
        \bottomrule
    \end{tabular}
    \begin{tablenotes}
        \item[1] Empty values exist because each author was evaluated with a distinct combination of perturbation types.
        \item[2] Refers to perturbation of contraction expansion and first-person pronoun transformations
    \end{tablenotes}
    \end{threeparttable}
\end{table}

\subsection{Visualization}

\begin{figure}[!htp]
     \centering
     \begin{subfigure}[t]{0.49\textwidth}
         \centering
         \includegraphics[width=\linewidth]{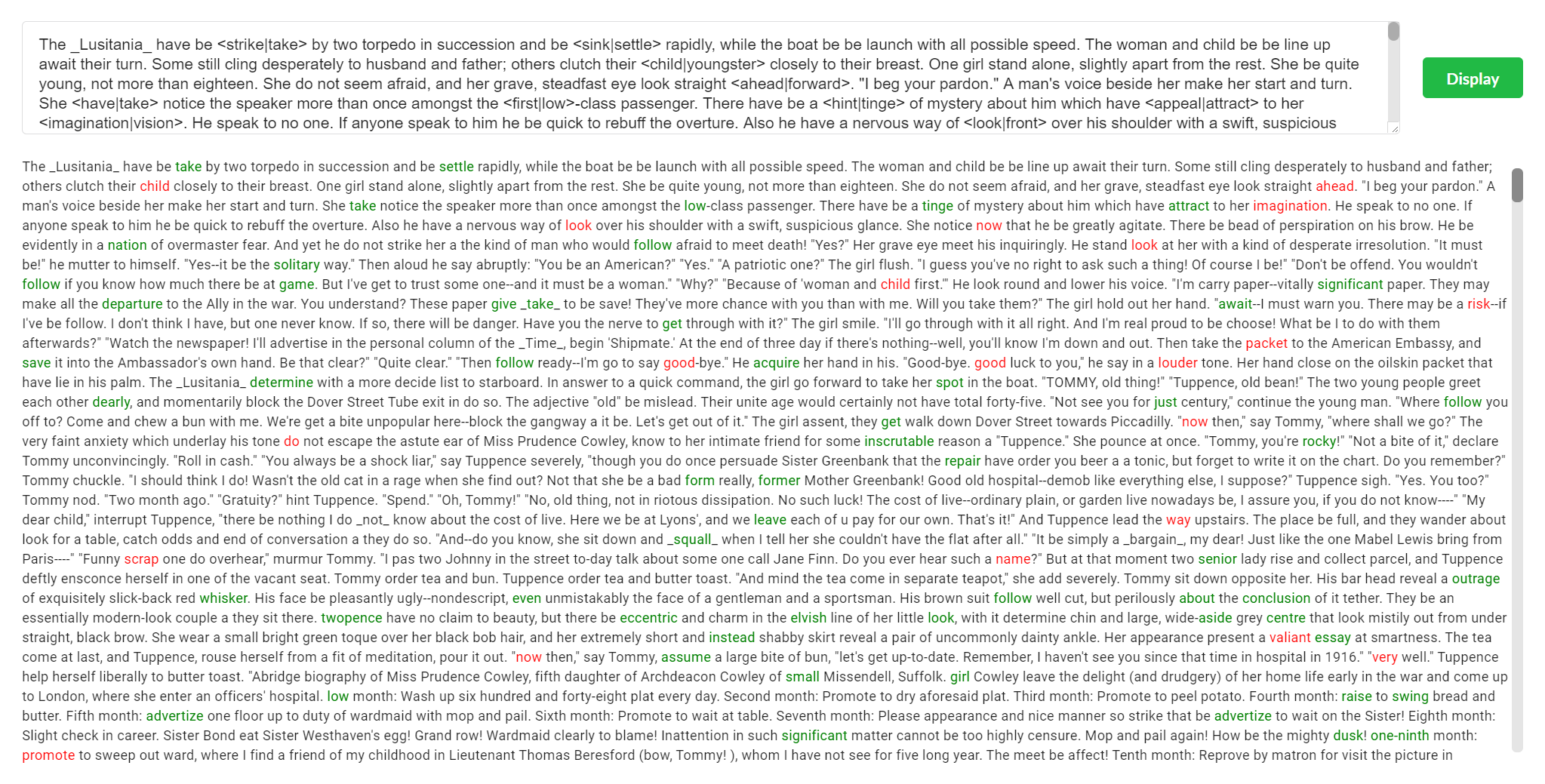}
         \caption {Web tool for validating text perturbations}
         \label{subfig:adversarial_helper}
     \end{subfigure}
     \hfill
     \begin{subfigure}[t]{0.49\textwidth}
         \centering
         \includegraphics[width=\linewidth]{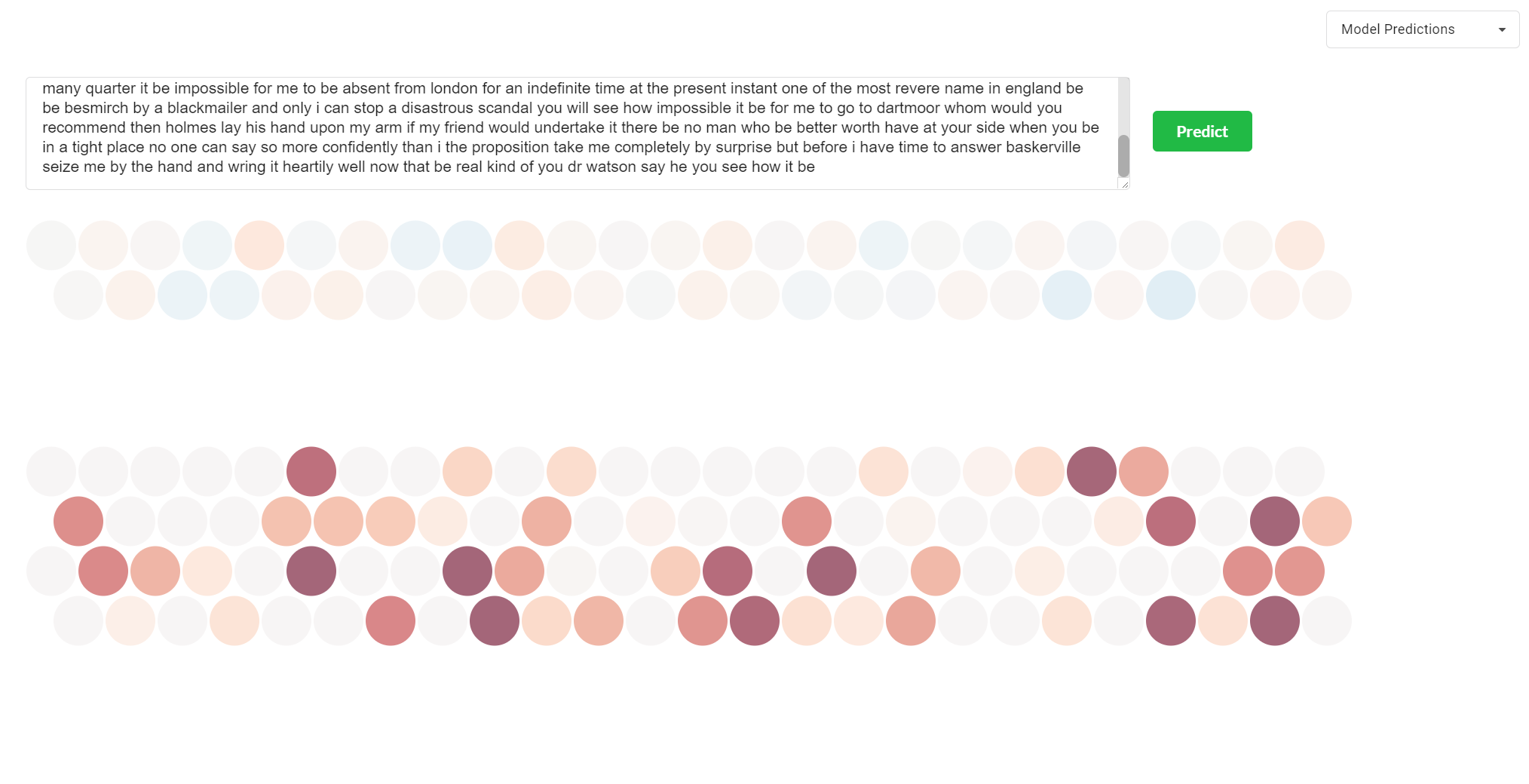}
         \caption {MLP architecture of author detection as a service}
         \label{subfig:crime_predictions}
     \end{subfigure}
     \par\bigskip
     \begin{subfigure}[t]{0.8\textwidth}
         \centering
         \includegraphics[width=\linewidth]{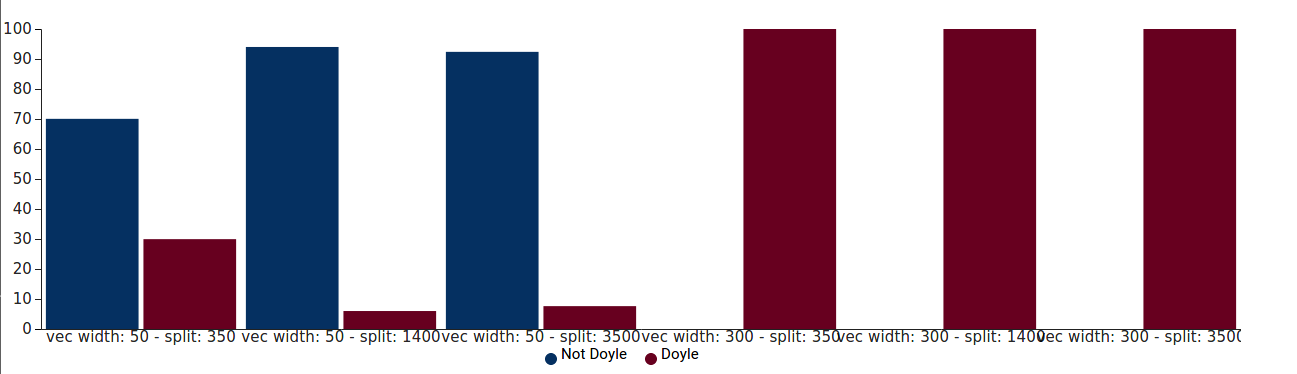}
         \caption {Prediction bars of author detection as a service}
         \label{subfig:prediction_bars}
     \end{subfigure}
     \caption{Web tools used for the author detection pipeline.}
     \label{fig:web_tools}
\end{figure}

In order to get better insight as to what the models were learning we built a web application using a popular web framework, \textit{React}\footnote{\url{https://reactjs.org}}, and \textit{D3.js}\footnote{\url{https://d3js.org}}. We visualized the MLP model's activations, weights, and outputs. We served the models and their predictions using \textit{Heroku}\footnote{\url{https://www.heroku.com}}, a company that provides platforms-as-a-service. The website (client) queries the Heroku server, providing a document to predict which is pre-processed based on \autoref{fig:preprocess}, runs the classification models, and records their activations, weights, and final predictions. The server then sends those values back to the client, which are then fed to our \textit{D3.js} visualization and used to update the node colors representative of their activations (\textcolor{BlueViolet}{blue} for negative, \textcolor{BrickRed}{red} for positive) and create a bar graph representing each model's final predictions, see \autoref{fig:web_tools}.

We also created another web capability that takes unperturbed text from the client, perturbs it on the server side, and then sends it back in a special format where perturbed words are replaced with \texttt{<original word|adversarial word>}. From there, the client displays this perturbed text and allows us to live-edit it, changing potentially perturbed words back and forth between their original and adversarial forms while keeping track of the perturbation ratio.

\section{Summary}

In this work, we explored the use of the author2vec algorithm for an authorship detection task based on three crime novelists: Agatha Christie, Sir Arthur Conan Doyle and Mary Roberts Rinehart. We used word2vec to create a word embedding representation for each author, and a simple MLP pipeline for the classification task. We achieved sufficiently high accuracy results using the author2vec algorithm. We also performed data perturbations on a subset of the three text corpora, exploring methods of synonym replacement, numbers to text, changing style of English between American and British, and finally, expanding contractions. We observed that we were able to retain significant classification accuracy for synonym replacement and language translation, but not with the other two forms of perturbations. The methods outlined in \Cref{futurework} might prove to have a significant impact on the performance of authorship detection tasks when the input corpora has been perturbed by adversarial input.

\section{Future Work}
\label{futurework}

The following three approaches can be considered as possible next steps in the authorship detection task. 
\begin{enumerate}[leftmargin=*]
    \item Feature Engineering

    In our approach to author detection, we only used the word2vec model to extract feature for each document. However, stylometric features, like lexical diversity, may also be discriminative in distinguishing authorship~\cite{iyer2019machine}. Among these stylometric features, POS tags may be easily obtained by using a pre-trained language model, and a sequence of POS tags can also be encoded as numerical feature and concatenated with word embedding. Since each author often has their unique sequence of POS tags associated with their writing styles, we envision combining POS tag and word2vec can further enhance the authorship detection. Moreover, document embeddings can be evaluated via the use of sub-word tokenization schemes, such as byte-pair encodings and unigram language models, and these could capture idiomatic and author-specific styles attributed to common orthographic syllables~\cite{kudo2018subword}. A writer's style can also be revealed by the distribution of word categorization and punctuation usage (i.e., formal vs. informal, complex vs. simple, rare vs. common). From our observations, a hypothesis we make is that the linguistic style of a literary writer is more pronounced in the non-conversational parts of the text given that characters' conversations tend to be sporadic, short, and may exhibit traits not necessarily associated with the author.
    Nevertheless, it may be that this is not the case and the writer's style dominates in conversational language.

    \item Data Augmentation

    Data augmentation has been widely used in supervised tasks, especially in the domain of computer vision. In this study, we implemented synonym replacement in the use of perturbation. On the other hand, synonym replacement can also be used for data augmentation in the domain of NLP to boost performance of text classification tasks~\cite{Wei_2019}. Instead of replacing synonyms defined by cosine similarity based on the trained word2vec model, we can also consider synonyms defined in English dictionary. Doing this will not confine us to corpus within the novels at hand, and it also expands the scope and diversity of input documents. Beside synonym replacement, a more aggressive approach for data augmentation in NLP task can be replacing paradigmatic relations~\cite{kobayashi-2018-contextual}. Kobayashi (2018) showed that paradigmatic relation ultimately changed the context while retaining sentiment and naturalness of text, and that this contextual augmentation also boosted performance of deep learning-based models. An extension to this work consists of evaluating the robustness of the document embedding and classification model with additional adversarial techniques which in some cases can be more disruptive than the ones presented. We note that applying perturbations that retain the original context and author style is a challenging task, and thus, these modifications may require manual evaluation. Examples include: tense change, point-of-view change, sentence reordering, to name a few.

    \item Classification Model

    Deep learning classification model has become popular choice to achieve state-of-art results. We implemented a straightforward MLP classification model in this task, but we envision adopting deeper neural networks will boost performance of authorship detection. A similar approach, combining word2vec and convolutional neural networks (CNN), was proposed for text classification~\cite{Helmy_2018}, and a deep model showed superior results compared to a shallower one~\cite{conneau2016deep}. We hypothesize that using word2vec and a deep CNN can be an alternate approach for authorship detection. On the other hand, transformer-based methods (BERT) have become the state-of-art method in various NLP tasks~\cite{vaswani2017attention}. A recent survey study showed that BERT outperformed traditional NN approaches in text classification across different data sets~\cite{jin2020bert}. Therefore, using the BERT model could be a promising approach to distinguishing authorship with state-of-art accuracy.
\end{enumerate}

\nocite{*}
\bibliographystyle{apacite}
\bibliography{main}


\end{document}